\documentclass[runningheads]{llncs}

\usepackage[T1]{fontenc}
\def\doi#1{\href{https://doi.org/\detokenize{#1}}{\url{https://doi.org/\detokenize{#1}}}}
\usepackage{graphicx, color, soul}
\usepackage{multirow}
\usepackage{listings}
\usepackage{caption}
\usepackage{subcaption}
\usepackage{algorithm}
\usepackage{algpseudocode}

\begin{document}
\title{Retrieval of surgical phase transitions using reinforcement learning}
\author{} 
\institute{}
%
%

	\author{Yitong Zhang\inst{1}\and 
		Sophia Bano\inst{1} \and
		Ann-Sophie Page\inst{2} \and
		Jan Deprest\inst{2} \and 
		Danail Stoyanov\inst{1} \and
		Francisco Vasconcelos\inst{1}
	}
	\institute{Wellcome/EPSRC Centre for Interventional and Surgical Sciences(WEISS) and Department of Computer Science, University College London, London, UK \and Department of Development and Regeneration, University Hospital Leuven, Leuven, Belgium
	}

\maketitle              
\begin{abstract}
In minimally invasive surgery, surgical workflow segmentation from video analysis is a well studied topic. The conventional approach defines it as a multi-class classification problem, where individual video frames are attributed a surgical phase label.
We introduce a novel reinforcement learning formulation for offline phase transition retrieval. Instead of attempting to classify every video frame, we identify the timestamp of each phase transition. By construction, our model does not produce spurious and noisy phase transitions, but contiguous phase blocks. We investigate two different configurations of this model. The first does not require processing all frames in a video (only $<60\%$ and $<20\%$ of frames in 2 different applications), while producing results slightly under the state-of-the-art accuracy. The second configuration processes all video frames, and outperforms the state-of-the art at a comparable computational cost. 
We compare our method against the recent top-performing frame-based approaches TeCNO and Trans-SVNet on the public dataset Cholec80 and also on an in-house dataset of laparoscopic sacrocolpopexy. We perform both a frame-based (accuracy, precision, recall and F1-score) and an event-based (event ratio) evaluation of our algorithms. 

\end{abstract}
\section{Introduction}

Surgical workflow analysis is an important component to standardise the timeline of a procedure. This is useful for quantifying surgical skills \cite{skill_analysis}, training progression \cite{intraoperative,surgical_instruction}, and can also provide contextual support for further computer analysis both offline for auditing and online for surgeon assistance and automation \cite{surgical_action_pred,activity_recognizing,generalizable_surgical_activity}. In the context of laparoscopy, where the main input is video, the current approaches for automated workflow analysis focus on frame-level multi-label classification. The majority of the state-of-the-art models can be decomposed into two components: feature extractor and feature classifier. The feature extractor normally is a Convolutional Neural Network (CNN) backbone converting images or batches of images (clips) into feature vectors. Most of the features extracted at this stage are spatial features or fine-level temporal features depending on the type of the input. Considering that long-term information in surgical video sequences aids the classification process, the following feature classifier predicts phases based on a temporally ordered sequence of extracted features. Following from natural language processing (NLP) and computer vision techniques, the architecture behind this feature classifier has evolved from Long Short-Term Memory (EndoNet, SVRCNet) \cite{EndoNet,SV-RCNet} , Temporal Convolution Network (TeCNO) \cite{tecno}, to Transformer (OperA, TransSV) \cite{OperA,Trans-SV} in workflow analysis. Although these techniques have improved over the years, the main problem formulation remains unchanged in that phase labels are assigned to individual units of frames or clips. 
\\
\\
These conventional models achieve now excellent performance on the popular Cholec80 benchmark \cite{EndoNet}, namely on frame-based evaluation metrics (accuracy, precision, recall, f1-score). However, small but frequent errors still occur throughout the classification of large videos, causing a high number of erroneous phase transitions, which make it very challenging to pinpoint exactly where one phase ends and another starts. To address this problem, we propose a novel methodology for surgical workflow segmentation. Rather than classifying individual frames in sequential order, we attempt to locate the phase transitions directly. Additionally, we employ reinforcement learning as a solution to this problem, since it has shown good capability in similar retrieval tasks \cite{joint_selection,bbox}. Our contributions can be summarised as follows:
\begin{itemize}
    \item We propose a novel formulation for surgical workflow segmentation based on phase transition retrieval. This strictly enforces that surgical phases are continuous temporal intervals, and immune to frame-level noise.
    \item We propose Transition Retrieval Network (TRN) that actively searches for phase transitions using multi-agent reinforcement learning. We describe a range of TRN configurations that provide different trade-offs between accuracy and amount of video processed.
    \item We validate our method both on the public benchmark Cholec80 and on an in-house dataset of laparoscopic sacropolpopexy, where we demonstrate a single phase detection application.
\end{itemize}

\section{Methods}

\begin{figure}[t]
\centering
\includegraphics[width=\textwidth]{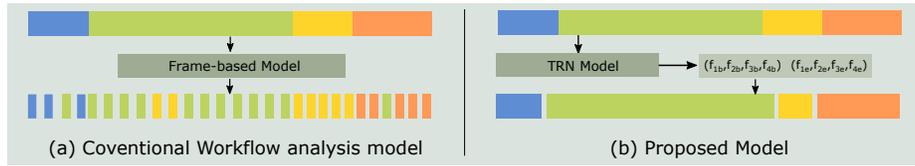}
\label{form}
\caption{Comparison of network architecture between (a) conventional model and (b) our proposed model with potential error illustration. The  conventional model assigns labels for each individual frames and our proposed model predicts frame indices for the starts and end position of phases.}
\label{fig:comparison}
\end{figure}

We consider the task of segmenting the temporal phases of a surgical procedure from recorded video frames. The main feature of our proposed formulation can be visualised in Fig. \ref{fig:comparison}. While previous work attempts to classify every frame of a video according to a surgical phase label, we attempt to predict the frame index of phase transitions. More specifically, for a surgical procedure with $N$ different phases, our goal is to predict the frame indices where each phase starts $\{f_{1b}, f_{2b} ... f_{Nb}\}$, and where each phase ends $\{f_{1e}, f_{2e} ... f_{Ne}\}$. If we can assume that surgical phases are continuous intervals, as it is often the case, then our approach enforces this by design. This is unlike previous frame-based approaches where spurious transitions are unavoidable with noisy predictions. To solve this problem we propose the Transition Retrieval Network (TRN), which we described next. 

\subsection{Transition Retrieval Network (TRN)}

\begin{figure}[t]
\centering
                \includegraphics[width=0.7\textwidth]{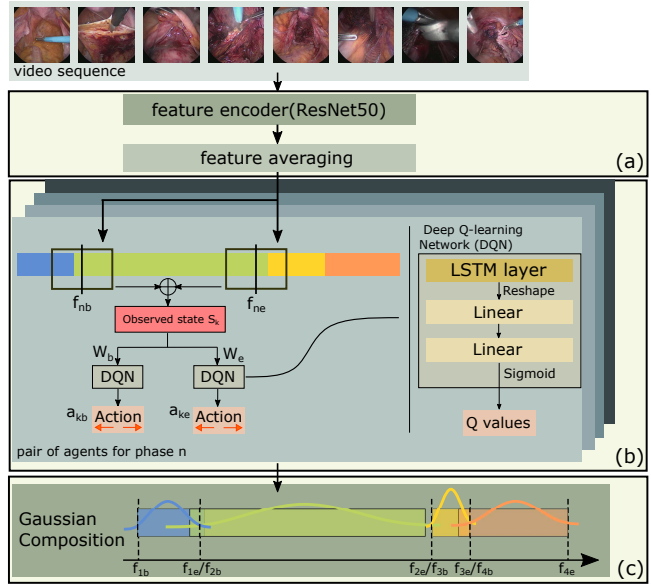}
\caption{TRN architecture with (a) averaged ResNet feature extractor, (b) multi-agent network for transition retrieval and (c) Gaussian composition operator}
\label{fig:architecture}
\end{figure}

Figure \ref{fig:architecture} shows the architecture of our TRN model. It has three main modules: an averaged ResNet feature extractor, a multi-agent network for transition retrieval, and a Gaussian composition operator to generate the final workflow segmentation result. 

\subsubsection{Averaged ResNet feature extractor} We first train a standard ResNet50 encoder (outputs 2048 dimension vector) \cite{ResNet} with supervised labels, in the same way as frame-based models. For a video clip of length $K$, features are averaged into a single vector. We use this to temporally down-sample the video through feature extraction. In this work we consider $K=16$.

\subsubsection{DQN Transition Retrieval} We first discuss the segmentation of a single phase $n$. We treat it as a reinforcement learning problem with 2 discrete agents $W_{b}$ and $W_{e}$, each being a Deep Q-Learning Network (DQN). These agents iteratively move a pair of search windows centered at frames $f_{nb}$ and $f_{ne}$, with length $L$. We enforced a temporal constraint where $f_{nb}{\leq}f_{ne}$ is always true. The state of the agents $s_{k}$ is represented by the $2L$ features within the search window, obtained with the averaged ResNet extractor. Based on their state, the agents generate actions $a_{kb}=W_{b}(s_k)$, and $a_{ke}=W_{e}(s_k)$, which move the search windows either one clip to the left or to the right within the entire video. During network training, we set a +1 reward for actions that move the search window center towards the groundtruth transition, and -1 otherwise. Therefore, we learn direction cues from image features inside the search windows. As our input to DQN is a sequence of feature vectors, a 3-layer LSTM \cite{LSTM} of dimension 2048 is introduced to DQN architecture for encoding the temporal features into action decision process. The LSTMs are followed by 2 fully connected layer (fc1 and fc2). Fc1 maps input search window of dimension $20L$ to $50$ and fc2 maps temporal features of dimension $50$ to the final 2 Q-values of 'Right' and 'Left'. We implemented the standard DQN training framework for our network. \cite{DQN_training} At inference time, we let the agents explore the video until they converge to a fixed position (i. e. cycling between left and right actions). Two important characteristics of this solution should be highlighted: 1) we do not need to extract clip features from the entire video, just enough for the agent to reach the desired transition; 2) the agents need to be initialised at a certain position in the video, which we discuss later. \looseness = -1



\subsubsection{Agent initialization configurations:}
We propose two different approaches to initialise the agents: fixed initialization (FI) and, ResNet modified initialization (RMI). FI initializes the search windows based on the statistical distribution (frame index average with relative position in video) of each phase transition on the entire training data. With FI, TRN can make predictions without viewing the entire video and save computation time. On the other hand, RMI initialises the search windows based on the averaged-feature ResNet-50 predictions by averaging the indices of all possible transitions to generate an estimation. In this way, we are very likely to have more accurate initialization positions to FI configuration and yield better performance.

\subsubsection{Merging different phases with Gaussian composition:} So far, we have only explained how our DQN transition retrieval model segments a single phase. To generalise this, we start by running an independently trained DQN transition retrieval model for each phase. If we take the raw estimations of these phase transitions, we inevitably create overlapping phases, or time intervals with no phase allocated, due to errors in estimation. To address this, we perform a Gaussian composition of the predicted phases. For each predicted pair of transitions $f_{nb}$, $f_{ne}$, we draw a Gaussian distribution centred at $\frac{f_{nb}+f_{ne}}{2}$, with standard deviation $\frac{|f_{nb}-f_{ne}|}{4}$. For each video clip, the final multi-class prediction corresponds to the phase with maximum distribution value.


\subsection{Training details}
The DQN model is trained in a multi-agent mode where $W_{b}$, $W_{e}$ for a single phase are trained together. The input for individual DQNs in each agent shares a public state concatenated from the content of both search windows, allowing the agents to be able to aware information of others. The procedures of training the DQN are showing in pseudo code in Algorithm \ref{alg:train}. For one episode, videos are trained one by one and the maximum number of steps an agent can explore in a video is 200 without early stopping. For every steps the agents made, movement information $(s_k, s_{k+1}, a_k, r_{k})$ are stored in its replay memories, and sampled with a batch size of 128 in computing Huber loss. \cite{DQN_training} This loss is optimized with gradient descent algorithm, where $\alpha$ is the learning rate and $\nabla_{W_{k}} \mathcal{L}_{k}$ is the gradient of loss in the direction of the network parameters.

\begin{algorithm}
\caption{The procedures of training DQN}\label{alg:train}
\begin{algorithmic}
\State Initialize parameters of agents $W_{b}$ and $W_{e}$ as $W_{0b}$ and $W_{0e}$
\State Initialize individual replay memories for agents $W_{b}$ and $W_{e}$
\For{$episode \gets 0$ to $episode_{MAX}$}
    \State Initialize search window positions (FI or RMI)
\For{$video \gets 0$ to $range(videos)$}
\For{$k \gets 0$ to $200$}
    \State $s_k \gets$ read ResNet features in search window
    \State $a_{kb} \gets W_{kb}(s_k)$ and $a_{ke} \gets W_{ke}(s_k)$ 
    \State $s_{k+1} \gets$ update search window position by $(a_{kb}, a_{ke})$ , read new features
    \State $r_{kb},r_{ke}  \gets$ compare $s_k$ and $s_{k+1}$ with reward function
    \State Save $(s_k, s_{k+1}, a_k, r_{kb})$ and $(s_k,s_{k+1}, a_k, r_{ke})$ into agent memory
    \State Compute loss $(\mathcal{L}_{kb}, \mathcal{L}_{ke})$ from random 128 samples from each memory
    \State Optimize $W_{kb}$: $W_{k+1b} \gets W_{kb}+\alpha \nabla_{W_{kb}} \mathcal{L}_{kb}$ 
    \State Optimize $W_{ke}$: $W_{k+1e} \gets W_{ke}+\alpha \nabla_{W_{ke}} \mathcal{L}_{ke}$
\EndFor
\EndFor
\EndFor
\end{algorithmic}
\end{algorithm}

\section{Experiment setup and Dataset Description}

The proposed network is implemented in PyTorch using a single Tesla V100-DGXS-32GB GPU of an NVIDIA DGX station. For the ResNet-50 part, PyTorch default ImageNet pretrained parameters are loaded for transfer learning and fine tuned for 100 epochs on Cholec80 and Sacrocolpopexy respectively. The videos are subsampled to 2.4 fps, centre cropped, and resized into resolution 224*224 to match the input requirement of ResNet-50. These ResNet parameters are shared for RMI, search windows, TecNO and TransSV in this research We train both ResNet-50 and DQN with Adam \cite{kingma2014adam} at a learning rate of 3e-4. For ResNet-50, we use a batch size of 100, where phases are sampled with equal probability. For DQN, the batch size is 128 sampled from a memory of size 10000 for each agent.

We tested the performance of TRN model on two datasets. Cholec80 is a publicly available benchamark that contains 80 videos of cholecystectomy surgeries \cite{EndoNet} divided into 7 phases. We use  40 videos for training, 20 for validation and 20 for testing. We also provide results on an in-house dataset of laparoscopic sacrocolpopexy \cite{sacro_implementation} containing 38 videos. It contains up to 8 phases (but only 5 in most cases), however, here we consider the simplified binary segmentation of the phases related to suturing a mesh implant (2 contiguous phases), given that suturing time is one of the most important indicators of the learning curve \cite{challenging_steps} in this procedure. We performed a 2-fold cross-validation with 20 videos for training, 8 for validation, and 10 for testing. For Sacrocolpopexy, we train our averaged ResNet extractor considering all phases, but train a single DQN transition retrieval for a suturing phase. We also do not require to apply Gaussian composition since we're interested in a single phase classification.


\subsubsection{Evaluation metrics:}
We utilise the commonly utilised frame-based metrics for surgical workflow: macro-averaged (per phase) precision and recall, F1-score calculated through this precision and recall, and micro-averaged accuracy. Additionally, we also provide event-based metrics that look at accuracy of phase transitions. An event is defined as block of consecutive and equal phase labels, with a start time and a stop time. We define event ratio as $\frac{E_{gt}}{E_{det}}$ where $E_{gt}$ is the number of ground truth events, and $E_{det}$ is the number of detected events by each method. We define a second ratio based on the Ward metric \cite{ward} which allocates events into sub-categories as deletion(D), insertion(I$'$), merge(M, M$'$), fragmentation(F, F$'$),  Fragmented and Merged(FM, FM$'$) and Correct(C) events. Here, we denote the Ward event ratio as ($\frac{C}{E_{gt}}$). For both of these ratios, values closer to 1 indicate better performance. Finally, whenever fixed initialisation (FI) is used, we also provide a coverage rate, indicating the average proportion of the videos that was processed to perform the segmentation. Lower values indicate fewer features need to be extracted and thus lower computation time.

\section{Results and Discussion}

We first provide an ablation of different configurations of our TRN model in Table \ref{tab:ablation}, for Cholec80. It includes two search window sizes (21 and 41 clips) and two initialisations (FI, RMI). The observations are straightforward. Larger windows induce generally better f1-scores, and RMI outperforms FI. This means that heavier configurations, requiring more computations, lead to better accuracies. Particular choice of a TRN configuration would depend on a trade-off analysis between computational efficiency and frame-level accuracy.

\begin{table}[t]
\centering
\resizebox{0.75\textwidth}{!}{%
\begin{tabular}{|l|l|l|l|l|l|l|l|l|}
\hline
Window size & Phase 1 & Phase 2 & Phase 3 & Phase 4 & Phase 5 & Phase 6 & Phase 7 & \begin{tabular}[c]{@{}l@{}}Overall \\ F1-score\end{tabular} \\ \hline
TRN21 FI  & 0.854 & 0.917 & 0.513 & 0.903 & 0.687 & 0.549 & 0.83  & 0.782 \\ \hline
TRN41 FI  & 0.828 & 0.943 & 0.636 & 0.922 & 0.558 & 0.694 & 0.85  & 0.808 \\ \hline
TRN21 RMI & 0.852 & 0.942 & 0.619 & 0.939 & 0.727 & 0.747 & 0.837 & 0.830 \\ \hline
TRN41 RMI & 0.828 & 0.940 & 0.678 & 0.945 & 0.753 & 0.738 & 0.861 & 0.846 \\ \hline
\end{tabular}%
}
\caption{TRN ablation in the Cholec80 dataset (F1-scores). The values per-phase are computed before Gaussian Composition, while the overall F1-score is for the complete TRN method.}
\label{tab:ablation}
\vspace{-6mm}
\end{table}

\begin{table}[t]
\centering
\resizebox{\textwidth}{!}{%
\begin{tabular}{|l|l|l|l|l|l|l|l|l|l|l|}
\hline
Dataset &
  Method &
  Accuracy &
  Precision &
  Recall &
  F1-Score &
  \begin{tabular}[c]{@{}l@{}}Event \\ ratio\end{tabular} &
  \begin{tabular}[c]{@{}l@{}}Ward \\ Event Ratio\end{tabular} &
  \begin{tabular}[c]{@{}l@{}}Coverage \\ rate(\%)\end{tabular} & 
  \begin{tabular}[c]{@{}l@{}}Computatio-\\ -nal Cost(s)\end{tabular}\\
  \hline
\multirow{6}{*}{Cholec80} & ResNet-50   & 79.7$\pm$7.5 & 73.5$\pm$8.4  & 78.5$\pm$8.9  & 0.756 & 0.120 & 0.375 & full & 96.6 \\ \cline{2-10} 
                          & TeCNO       & 88.3$\pm$6.5 & 78.6$\pm$9.9  & 76.7$\pm$12.5 & 0.774 & 0.381 & 0.691 & full & 99.6  \\ \cline{2-10} 
                          & Trans-SVNet & 89.1$\pm$5.7 & 81.7$\pm$6.5  & 79.1$\pm$12.6 & 0.800 & 0.316 & 0.566 & full & 99.6  \\ \cline{2-10} 
                          & TRN21 FI    & 85.3$\pm$9.6 & 78.1$\pm$11.1 & 78.9$\pm$13.5 & 0.782 & 1     & 0.934 & 57.6 & 60.6 \\ \cline{2-10} 
                          & TRN41 FI    & 87.8$\pm$8.1 & 80.3$\pm$9.1  & 81.7$\pm$12.4 & 0.808 & 1     & 0.956 & 59.1 & 64.9 \\ \cline{2-10} 
                          & TRN41 RMI   & 90.1$\pm$5.7 & 84.5$\pm$5.9  & 85.1$\pm$8.2  & 0.846 & 1     & 0.985 & full & 105.5 \\ \hline
                          \hline
\multirow{5}{*}{\begin{tabular}[c]{@{}l@{}}Sacrocol-\\ -popexy\end{tabular}} &
  ResNet-50 &
  92.5$\pm$3.8 &
  94.9$\pm$2.8 &
  84.5$\pm$8.4 &
  0.892 &
  0.029 &
  0.016 &
  full & 493.7 \\ \cline{2-10} 
                          & TeCNO       & 98.1$\pm$1.7 & 97.7$\pm$1.9  & 97.5$\pm$3.0  & 0.976 & 0.136 & 0.438 & full & 493.8 \\ \cline{2-10} 
                          & Trans-SVNet & 97.8$\pm$2.2 & 96.5$\pm$4.5  & 98.0$\pm$3.5  & 0.971 & 0.536 & 0.813 & full & 493.9 \\ \cline{2-10} 
                          & TRN21 FI    & 89.8$\pm$6.2 & 88.6$\pm$11.7 & 85.3$\pm$11.1 & 0.860 & 0.971 & 0.875 & 14.6 & 78.1\\ \cline{2-10} 
                          & TRN81 FI    & 90.7$\pm$6.1 & 88.6$\pm$11.5 & 88.5$\pm$11.1 & 0.875 & 0.941 & 0.860 & 18.3 & 104.0\\ \hline
\end{tabular}%
}
\caption{Evaluation metric results summary of ResNet-50, our implementation of TeCNO and Trans-SV, and ablative selected TRN result on Cholec80 and Sacrocolpopexy. The computatinla cost is in average second to process a single video}
\label{tab:results}
\vspace{-7mm}
\end{table}

\begin{figure}[t]
     \centering
     \begin{subfigure}[b]{\textwidth}
         \centering
         \includegraphics[width=0.95\textwidth]{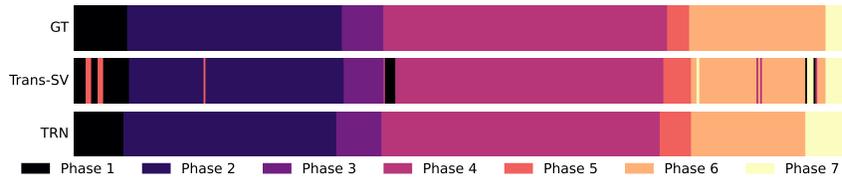}
         \caption{An example of video77 from Cholec80 processed by Trans-SV and TRN41 RMI}
         \label{fig:cholecexample}
     \end{subfigure}
     \hfill
     \begin{subfigure}[b]{\textwidth}
         \centering
         \includegraphics[width=0.9\textwidth]{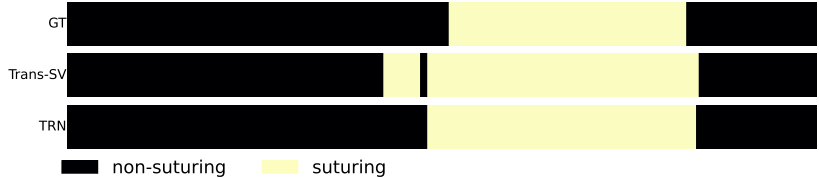}
         \caption{An example video from Sacrocolpopexy processed by Trans-SV and TRN81 FI}
         \label{fig:three sin x}
     \end{subfigure}
        \caption{Color-coded ribbon illustration for two complete surgical videos from (a) Cholec80 and (b) Sacrocolpopexy processed by Trans-SV and TRN models.}
        \label{fig:example_seq}
\end{figure}

Table \ref{tab:results} shows a comparison between TRN and state-of-the-art frame-based methods on both Cholec80 and Sacrocolpopexy. The utilised baselines are TeCNO \cite{tecno}, Trans-SVNet \cite{Trans-SV}, which we implemented and trained ourselves. Instead of simple ResNet50, we use the same feature averaging process as the TRN for consistency. Also for consistency, we disabled causal convolution in TCN (it is a provided flag in their code) that allowing Trans-SV and TCN to be trained in off-line mode .\\
\\
For Cholec80, our full-coverage model (TRN41 RMI) surpasses the best baseline (Trans-SVNet) in all frame-based metrics, while having significantly better even-based metrics (event ratio, Ward event ratio). This can be explained by TRN's immunity to frame-level noisy predictions, which can be visualised on a sample test video in Fig. \ref{fig:cholecexample}. Remaining visualisations for all test data are provided in supplementary material. 

Still for Cholec80, our partial-coverage models (TRN21/41 FI) have frame-based metrics below the state-of-the-art baselines, however, they have the advantage of performing segmentation by only processing below 60\% of the video samples. The trade-off between coverage and accuracy can be observed. Additionally, TRN21/41 FI also have substantially better event-based metrics than frame-based methods due to its formulation.
An operation may not have a complete set of phases. For the missing phases in Cholec80 test videos, it has little impact as shown in supplementary document. The RL agent makes begin and end labels converge towards the same/consecutive timestamps. Sometimes there are still residual frames erroneously predicted as the missing phase. These errors are counted in reported statistics resulted in imperfect event ratios


For sacrocolpopexy, we display a case where our partial-coverage models (TRN21/81 FI) are at their best in terms of computational efficiency. These are very long procedures and we are interested in only the suturing phases as an example of clinical interest. \cite{lamblin2021glue} Therefore, a huge proportion of the video can be ignored for a full segmentation. Our models slightly under perform all baselines in frame-based  metrics, but achieve this result by only looking at under 20\% of the videos on average.

\section{Conclusion}

In this work we propose a new formulation for surgical workflow analysis based on phase transition retrieval (instead of frame-based classification), and a new solution to this problem based on multi-agent reinforcement learning (TRN). This poses a number of advantages when compared to the conventional frame-based methods. Firstly, we avoid any frame-level noise in predictions, strictly enforcing phases to be continuous blocks. This can be useful in practice if, for example, we are interested in time-stamping phase transitions, or in detecting unusual surgical workflows (phases occur in a non-standard order), both of which are challenging to obtain from noisy frame-based classifications. In addition, our models with partial coverage (TRN21/41/81 FI) are able to significantly reduce the number of frames necessary to produce a complete segmentation result.

There are, however, some limitations. First, there may be scenarios where phases occur with an unknown number of repetitions, which would render our formulation unsuitable. Our TRN method is not suitable for real-time application, since it requires navigating the video in arbitrary temporal order. TRN may have scalability issues, since we need to train a different agent for each phase, which may be impractical if a very large number of phases is considered. This could potentially be alleviated by expanding the multi-agent framework to handle multiple phase transitions simultaneously. TRN is also sensitive to agent initialisation, and while we propose 2 working strategies (FI, RMI), they can potentially be further optimised.



\subsubsection{Acknowledgements} 
This research was supported by the Wellcome/EPSRC Centre for Interventional and Surgical Sciences (WEISS) [203145/Z/16/Z]; the Engineering and Physical Sciences Research Council (EPSRC) [EP/P027938/1, EP/R004080/1, EP/P012841/1]; the Royal Academy of Engineering Chair in Emerging Technologies Scheme, and Horizon 2020 FET Open (863146).

%
%
%
\bibliographystyle{splncs04}
\bibliography{ref}

\end{document}